\documentclass[10pt,twocolumn,letterpaper]{article}

\usepackage{iccv}              

\definecolor{iccvblue}{rgb}{0.21,0.49,0.74}
\usepackage[pagebackref,breaklinks,colorlinks,allcolors=iccvblue]{hyperref}

\usepackage{multirow}
\usepackage{xcolor,colortbl}

\def\paperID{} 
\def\confName{ICCV}
\def\confYear{2025}

\title{Probabilistic Modeling of Disparity Uncertainty for Robust and Efficient Stereo Matching} 

\author{Wenxiao Cai$^{1,3}$,\hspace{0.7em}Dongting Hu$^{2}$,\hspace{0.7em} Ruoyan Yin$^{4}$,\hspace{0.7em}Jiankang Deng$^{5}$ \\ Huan Fu$^{6}$,\hspace{0.7em}Wankou Yang$^{1}$\thanks{Corresponding author: wkyang@seu.edu.cn},\hspace{0.7em}Mingming Gong$^{2}$ \\
$^1$ Southeast University\hspace{0.7em}
$^2$ The University of Melbourne\hspace{0.7em}
$^3$ Stanford University\\
$^4$ National University of Singapore\hspace{0.7em}
$^5$ Imperial College London\hspace{0.7em}
$^6$ Alibaba Group
}

\begin{document}
\maketitle

\begin{abstract}
   Stereo matching plays a crucial role in various applications, where understanding uncertainty can enhance both safety and reliability.
   Despite this, the estimation and analysis of uncertainty in stereo matching have been largely overlooked.
   Previous works struggle to separate it into data (aleatoric) and model (epistemic) components and often provide limited interpretations of uncertainty. 
   This interpretability is essential, as it allows for a clearer understanding of the underlying sources of error, enhancing both prediction confidence and decision-making processes.
   In this paper, we propose a new uncertainty-aware stereo matching framework.
   We adopt Bayes risk as the measurement of uncertainty and use it to separately estimate data and model uncertainty. 
   We systematically analyze data uncertainty based on the probabilistic distribution of disparity and efficiently estimate model uncertainty without repeated model training.
   Experiments are conducted on four stereo benchmarks, and the results demonstrate that our method can estimate uncertainty accurately and efficiently, without sacrificing the disparity prediction accuracy.
\end{abstract}

\section{Introduction}
\label{sec:intro}
Stereo matching, or binocular depth estimation, has long been a meaningful problem with numerous important applications in autonomous driving~\cite{autodrive1,autodrive2}, embodied AI~\cite{robot1,robot2}, industrial quality control~\cite{quality} and medical image processing~\cite{medical1,medical2}. 
Uncertainty estimation in stereo matching is crucial and has many real-world applications. In autonomous driving, for example, the accuracy of depth estimation may decrease on foggy days. If the car is aware that uncertainty in the estimation is high under such circumstances, it can make more informed decisions, such as slowing down to navigate more carefully~\cite{over-conf1,over-conf2}.
Therefore, it is crucial to recognize uncertainties in stereo matching.

\begin{figure*}[t]
	\begin{center}
    \includegraphics[width=1\linewidth]{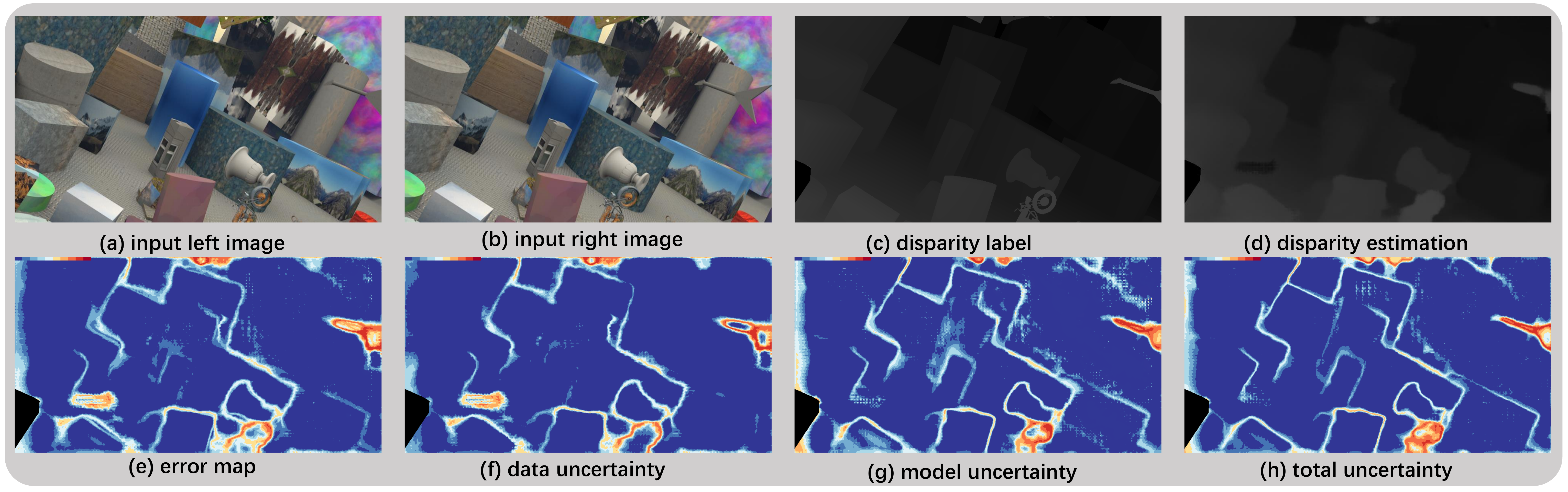}
	\end{center}
 	\caption{(a)(b): inputs of stereo matching. (c)(d): disparity ground truth and prediction. (e): disparity prediction error. (f)(g)(h): Our estimation of data, model and total uncertainty. The estimated uncertainty aligns well with prediction error.}
	\label{fig:teaser}
\end{figure*}

Uncertainty can be classified into two main categories: data uncertainty and model uncertainty~\cite{def-uncert}. Data (aleatoric) uncertainty arises from the inherent uncertainty in the data generation process, while model (epistemic) uncertainty stems from the inadequacy of the fitted model and the stochastic nature of model parameters, which can be reduced by increasing the amount of training observation~\cite{whatuncert}. 
Separately estimating these uncertainties provides several key advantages:
\begin{itemize}
    \item Data uncertainty can help filter high-quality data: Points with high data uncertainty may result from errors in the data collection process and might not reflect the true ground truth. Identifying such points allows us to filter low-quality data, leading to a cleaner, more robust dataset that better supports accurate model training.
    \item Model uncertainty guides data collection: Separate estimates of model uncertainty can highlight data regions where the model is most uncertain, which often indicates a need for more training samples. 
\end{itemize}

Estimating data and model uncertainty in stereo matching presents several challenges:
(i) Existing frameworks~\cite{ucfnet,elfnet} primarily focus on estimating total uncertainty without disentangling it into its constituent sources, namely data and model uncertainty.
(ii) Robustly estimating data uncertainty requires probabilistic modeling of the disparity distribution, a task that remains unclear within current learning frameworks for stereo matching~\cite{sednet}. 
(iii) Estimating model uncertainty often requires multiple re-trainings, such as deep ensembles~\cite{ensemble1,ensemble2}, leading to slower processing speeds and substantial computational demands.

We address these issues by probabilistically modeling distribution of disparity.
Bayes risks of the estimated disparity distribution is adopted as a measurement of uncertainty. 
We introduce a framework that estimates the probability distribution of disparity values with ordinal regression~\cite{or}, which models the distribution better than simply supervising on the ground truth value.
Moreover, we estimate model uncertainty through an additional kernel-based estimator applied to model embeddings. 
This approach efficiently estimates data and model uncertainty separately, without requiring model retraining.

Experimental results on four widely used stereo matching datasets~\cite{kitti,vk2,sceneflow,drivingstereo} show that our proposed method has a more reliable estimation of data and model uncertainty. 
Our uncertainty aware model also shows improved performance in disparity estimation.
Moreover, we demonstrate that prediction accuracy can be further improved by selecting data points with small uncertainties.
The contributions of our paper are:
\begin{itemize}
\item We build an uncertainty-aware stereo matching framework on Bayes risk.
\item Our method can robustly estimate the data uncertainty by probabilistically modeling disparity distribution.
\item We adopt an extra kernel regression model to efficiently estimate model uncertainty.
\item We further illustrate the potentials of our estimated uncertainty in enhancing prediction accuracy by effectively filtering out highly uncertain data.
\end{itemize}

\section{Related Work}
In this section, we introduce recent works in stereo matching, depth estimation, and uncertainty estimation. The calculation and application of uncertainty in stereo matching is also explored.

\subsection{Uncertainty Quantification}
Uncertainty is an important topic, and there are many efforts attempting to quantify and decouple data and model uncertainty.
~\cite{whatuncert} proposed the use of a Bayesian deep learning framework for estimating uncertainty, considering both data and model uncertainty. Most methods utilize MC Dropout~\cite{mcdropout1,mcdropout2,mcdropout3}, Deep Ensemble~\cite{ensemble1,ensemble2}, and Variational Inference (VI)~\cite{vi1,vi2,vi3} to compute model uncertainty, resulting in prolonged computational time. Deep Evidential Regression~\cite{deep-evidential-regression} suggests that uncertainty arises from a higher-dimensional distribution, thereby avoiding the time and computational costs associated with repeated model training. NUQ~\cite{nuq,duq,ddu}, through a posterior approach, trains Kernel Density Estimation (KDE) or Gaussian Mixture Model (GMM) to estimate model and data uncertainty. 

\subsection{Stereo Matching and Depth Estimation}
Stereo matching, disparity estimation, and binocular depth estimation are similar concepts because disparity and depth can be converted into each other: $ disparity = f*B/depth$, where $f$ is the focal length, and $B$ is the baseline distance between two cameras. 
Typically, features from left and right images are used to build cost volumes with cross-correlation or concatenation, followed by a 2D or 3D convolutional neural network~\cite{pcwnet,disparity1,disparity2,disparity3,disparity4,disparity5,disparity6}.
GwcNet~\cite{gwcnet} is a state-of-the-art model which constructs the cost volume by group-wise correlation. We use GwcNet as the backbone for depth estimation.

\begin{figure*}[t!]
	\begin{center}
    \includegraphics[width=1\linewidth]{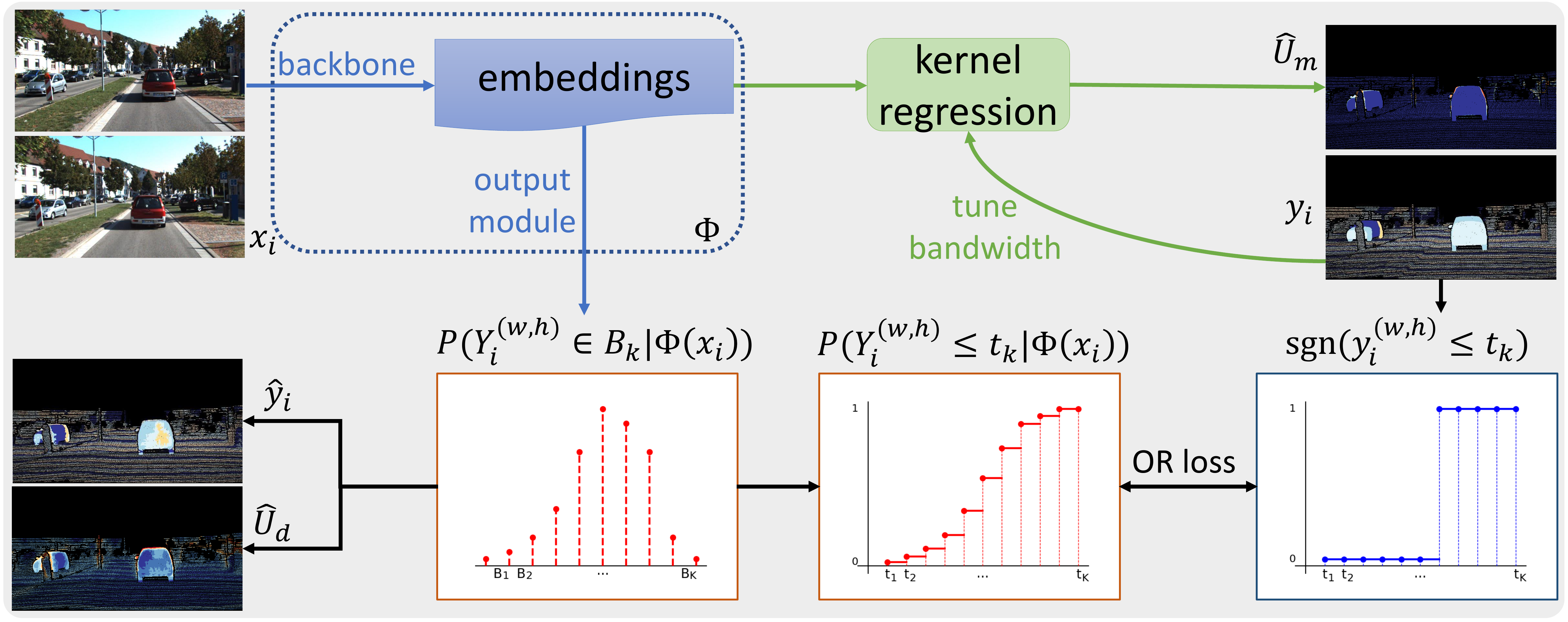}
	\end{center}
 	\caption{Pipeline of the proposed stereo matching and uncertainty quantification method. Ordinal regression model is adopted to estimate PMF of disparity. Prediction is the expectation of PMF, and data uncertainty is PMF's variance. The model is supervised by the ordinal label, using ordinal regression loss. Model uncertainty is estimated with a kernel regression model, which is fitted on the embeddings of OR model. }
	\label{fig:pipeline}
\end{figure*}

\subsection{Uncertainty in Stereo Matching}
ELFNet~\cite{elfnet} and UCFNet~\cite{ucfnet} employ non-deep-learning methods to calculate uncertainty in stereo depth estimation~\cite{DeepSU,GradientbasedUF}, which is then utilized in downstream tasks such as transfer learning. SEDNet~\cite{sednet} incorporates an uncertainty module after the depth estimation network, employing KL divergence to have the uncertainty module learn the distribution of depth estimation errors as uncertainty.
Deep distribution regression~\cite{ddr} was adopted in monocular depth estimation to estimate data uncertainty~\cite{conor}.
In this paper,  we use ordinal distribution regression to learn the distribution of disparity and compute the data uncertainty by its variance.
A time-saving but accurate method is adopted to estimate model uncertainty.

\section{Proposed Method}
We use Bayes risk for decomposition because it directly quantifies the model's risk by measuring the discrepancy between predictions and ground truth. This decomposition is suitable for uncertainty quantification in stereo matching as it effectively captures uncertainties arising from data noise and model error in depth estimation.
In this section, we first show the definition of uncertainty as Bayes risks~\cite{nuq}. Then we illustrate our data-driven uncertainty estimation method in stereo matching.
We use Ordinal Regression (OR)~\cite{or} to estimate disparity (or depth) distribution. Model uncertainty is calculated from parameters of an extra Kernel Regression (KR), fitted on the embedding of OR model. 
The pipeline is shown in Fig. \ref{fig:pipeline}.

\subsection{Definition of Uncertainty}
Let two random variables $X$ and $Y$ denote the input stereo images and the disparity, respectively. Consider a training set $D_{train}=\{x_i,y_i\}_{i=1}^N$ drawn from an unknown joint distribution $D$, the optimal rule $g^*$ can be learned by minimizing the expected risk $R(g)=E_{(X,Y)\sim D}l(g(X),Y)$, where $l$ is a loss function, and the rule $\hat{g}$ trained on observations from $D$ is the minimizer of the empirical risk $R_N(g)=\frac{1}{N}\sum_{i=1}^n l(g(x_i),y_i)$.
For reader's convenience of understanding, we first show total and data uncertainty ($U_t, U_d$) definitions in classification problems:

\begin{equation}
U_t = P(\hat{g}(X) \neq Y | X=x), U_d = P(g^*(X) \neq Y | X=x). 
\end{equation}
Data uncertainty is the Bayes risk of the best model $g^*$ fitted on the whole data distribution $D$.
Total uncertainty measures the probability of error, of model $\hat{g}$. As model uncertainty reflects the imperfectness of fitted model $\hat{g}$ in comparasion to $g^*$, we can define it as:
\begin{equation}
    U_m = U_t - U_d.
\end{equation}

In the regression problem of stereo depth estimation, a similar definition is used~\cite{nuq}:
\begin{equation}
\begin{aligned}
&U_t = E\left[(\hat{g}(X) - Y)^2 | X=x)\right],\\ 
&U_d = E\left[(g^*(X) - Y)^2 | X=x)\right],\\ 
&U_m = U_t - U_d.
\end{aligned}
\label{eq:regression-uncert}
\end{equation}
Since $g^*$ is fitted on whole distribution $D$, it should be: $g^*(X)=E[Y|X]$. 
As a result, $U_d = E\left[(g^*(X) - Y)^2 | X=x)\right] = Var\left[Y\right|X=x]$.
From this definition, it is evident that data uncertainty is an inherent characteristic of the data itself and is independent of the model or method used for estimation.


It is essential to note that while uncertainty and error are closely related, they are not the same concept. Error typically refers to the difference between the model's prediction and the true value, whereas uncertainty focuses on error in the whole distribution $D$. 
For long-tail distribution or OOD data, a model have a chance to luckily exhibit a small error but it should always detect a high uncertainty. 
Additionally, on a test set, uncertainty can be predicted, whereas the model cannot predict error on test set (or the prediction accuracy should be 100$\%$).

\subsection{Stereo Matching and Data Uncertainty}
In data-driven uncertainty estimation, we estimate uncertainty $U_d$ on observation dataset. Estimated data uncertainty $\hat{U}_d$ is obtained by estimating distribution of $Y$:
\begin{align}
U_d = Var\left[Y|X\right], \hat{U}_d = Var\left[\hat{Y}|X\right].
\end{align}
We propose to use an ordinal regression (OR) model to estimate disparity distribution in stereo matching.
Thus the estimation of disparity $\hat{Y}_i$ and estimation of data uncertainty $\hat{U}_d$ can be obtained by the mean and variance of estimated distribution.
For disparity within the range [$\alpha, \beta$], we divide it into $K$ bins. $B_k = (t_{k-1},t_k]$, where:

\begin{equation}
    t_k = \alpha + k (\beta - \alpha) / K, \; for \, k \in \{ 0,1,...,K \}.
\end{equation}

For disparity estimation, we choose the bins so that each bin corresponds to a disparity value.
OR tries to predict the probability of a pixel's disparity falling into $B_k$.
For an image $x_i$ of width $W$, height $H$, and channel $C$, and OR model $\Phi$, the feature map $\eta_i = \Phi (x_i)$ is of size $(W,H,K)$. Here $K$ is the features dimension, and also the number of bins.
For a pixel at position $(w,h)$, the $K$ values output by the softmax layer represent the probabilities of disparity $Y_i^{(w,h)}$ falling into the $K$th bin, i.e., the conditional probability mass function (PMF):

\begin{align}
\begin{aligned}[t]
    P(Y_i^{(w,h)} \in B_k | \Phi(x_i)) &= \frac{e^{\eta_{i,k}^{(w,h)}}}{\sum_{j=1}^{K}e^{\eta_{i,k}^{(w,h)}}}, \; \\ &for \, k \in \{1,2,...,K\}.
\end{aligned}
\end{align}

We recast the stereo matching problem to a series of binary classification tasks answers: what is the probability of the disparity value being less or equal to $t_k$?
To do so, we first convert the PMF to the conditional cumulative distribution function (CDF) by summing PMF values:
\begin{align}
P(Y_i^{(w,h)} \leq t_k | \Phi(x_i)) &=  \sum_{j=1}^{k} P(Y_i^{(w,h)} \in B_k | \Phi(x_i)).
\end{align}

CDF is the likehood of a pixel's disparity value less or equal to threshold $t_k$.
To estimate the likelihood of a pixel's disparity less or equal to $t_k$, we adopt binary classification loss for each $t_k$, and sum them up as the OR loss.
For ground truth disparity image $y_i$, loss $l$ of a pixel and loss $L$ for an image are expressed as:

\begin{align}
    \mathcal{P}_{i}^{(w,h)} &= \log(P(Y_i^{(w,h)} \leq t_k | \Phi(x_i))), \\
    l(x_i,y_{i}^{(w,h)},\Phi) &= \begin{aligned}[t] & -\sum_{k=1}^{K} 
        \left[sgn(y_{i}^{(w,h)} \leq t_{k}) \mathcal{P}_{i}^{(w,h)} \right. \\
     & \left.+(1-sgn(y_{i}^{w,h} \leq t_{k}) (1-\mathcal{P}_{i}^{(w,h)})\right] ,
    \end{aligned} \\
    L(x_i,y_i,\Phi) &= \frac{1}{WH} \sum_{w=0}^{W-1} \sum_{h=0}^{H-1} l(x_i,y_{i}^{(w,h)}).
\end{align}
The estimator $\hat{\Phi} = argmin_{\Phi} \sum_{i=1}^{N} l(x_i,y_i,\Phi)$ is trained on dataset $D_{train}$, which consist of $N$ pixels. 
 We consider the expectation of conditional PMF as our disparity estimation $\hat{y}^{(w,h)}$.
 Since the variance of $Y$ distribution is its data uncertainty, we consider variance of estimated conditional PMF as estimated data uncertainty $\hat{U}_d$:

\begin{align}
    \hat{y}^{(w,h)} &= \frac{1}{K} \sum_{k=1}^{K} \frac{t_k + t_{k+1}}{2} P(Y_i^{w,h} \in B_k | \Phi(x_i)), \\
    \hat{U}_{d}^{(w,h)} &=  \begin{aligned}[t] \sum_{k=1}^{K} (\frac{t_k + t_{k+1}}{2} - E\left[\hat{y}^{(w,h)}|x;\hat{\Phi}\right]) \\
    P(Y_i^{(w,h)} \in B_k | \Phi(x_i)).  \end{aligned}
\end{align}

Intuitively, variance of distribution reflects the spread of predicted probability mass. If the variance is small, (e.g. 99\% for one bin $B_{11}$ and 1\% for another one), the model is sure that the disparity value should fall in $B_{11}$ and uncertainty should be small. If the variance is larger, (e.g. 33\% for bin $B_{10}$, $B_{11}$ and $B_{12}$), the model is unsure of which bin the disparity should fall into. Thus uncertainty should be larger. However, for each case, the expectation of distribution should be somewhere around $B_{11}$. So the predicted disparity should fall into $B_{11}$ with different uncertainties.

\subsection{Model Uncertainty}
Model uncertainty is represented by the excess risk between total and data uncertainty.
$g^*$ is fitted on the whole distribution $D$ and is optimal on $D$'s observation sets $D_{train}$ and $D_{test}$.
$\hat{g}$ is fitted on $D_{train}$, as an approximation of $g^*$.
According to Eq.~\ref{eq:regression-uncert}, data uncertainty $U_d$ corresponds to Bayes risk of $g^*$: $U_d = E\left[(g^*(X) - Y)^2 | X=x)\right]$,
and total uncertainty $U_t$ corresponds to Bayes risk of  $\hat{g}$: $U_t = E\left[(\hat{g}(X) - Y)^2 | X=x)\right]$. 
The excess risk $E\left[(\hat{g}(X) - Y)^2 | X=x)\right] - E\left[(g^*(X) - Y)^2 | X=x)\right]$ can be seen as the imperfectness of model $\hat{g}$. Thus it corresponds to model uncertainty $U_m$.
Let's consider a particular estimator $g$ first and then find the difference between $g^*$ and $\hat{g}$.

We choose Kernel Regression (KR) model because it allows for a simple mathematical description of difference between $g^*$ and $\hat{g}$~\cite{converge}.
We train KR model on the embeddings of the OR model $\eta_{i}^{(w,h)}$ and corresponding disparity label: $Y_i^{(w,h)}$ ($i\in\{1,...,N\}$,$w\in\{1,...W\}$, $h\in\{1,...,H\}$). 
To express it clearly, we denote the KR dataset in this section as ${\{s_i,t_i\}}_{i=1}^{M}$ ($M=H*W*N$, $s_i$ in feature and $t_i$ is its label).
Let $K_h$ denote the kernel function, kernel regression estimator $\hat{g}$ is defined as:

\begin{equation}
    \hat{g}(s) = \frac{\sum_{i=1}^{M} K_{h} (s-s_{i})t_i}{\sum_{i=1}^{M}K_{h} (s-s_{i})}. 
\end{equation}

According to~\cite{converge}, the difference between $g^*$ and $\hat{g}$ converges in a Gaussian distribution:

\begin{equation}
    g^*(x) - \hat{g}(x) \rightarrow \frac{C}{N} \frac{\sigma^{2}(x)}{p(x)},
\end{equation}
where $C$ is a constant related to kernel and bandwidth, $N$ is the number of data in observation set, $\sigma^{2}$ is the variance of $x$ in the observation set, and $p(x)$ is the marginal distribution of covariates. 
An asymptotic approximation of $g^*(x) - \hat{g}(x)$ is used in estimating $U_m$: $U_m = 2 \sqrt{\frac{2}{\pi} \frac{C}{N} \frac{\sigma^{2}(x)}{p(x)}}$ (details in~\cite{nuq}).

In kernel regression of our work, an RBF Kernel is adopted:
\begin{equation}
    K_h(x,x_i) = exp(\frac{-(s-s_i)^2}{2\sigma^2}),
\end{equation}
where $\sigma$ is the bandwidth, and $K_h$ is the sum of all dimensions of $s$. 
K-Nearest Neighbours(KNN) is adopted in KR to further reduce computational costs. KR is a post-process fashion and does not require retraining the whole OR model, making it faster compared to methods such as Deep Ensemble~\cite{ensemble1,ensemble2} and Bootstraping like Wild and Multiplier Bootstrap~\cite{wildbs,multiplierbs}.

\section{Experiments}

\begin{figure*}[h!]
	\begin{center}
		\includegraphics[width=1\linewidth]{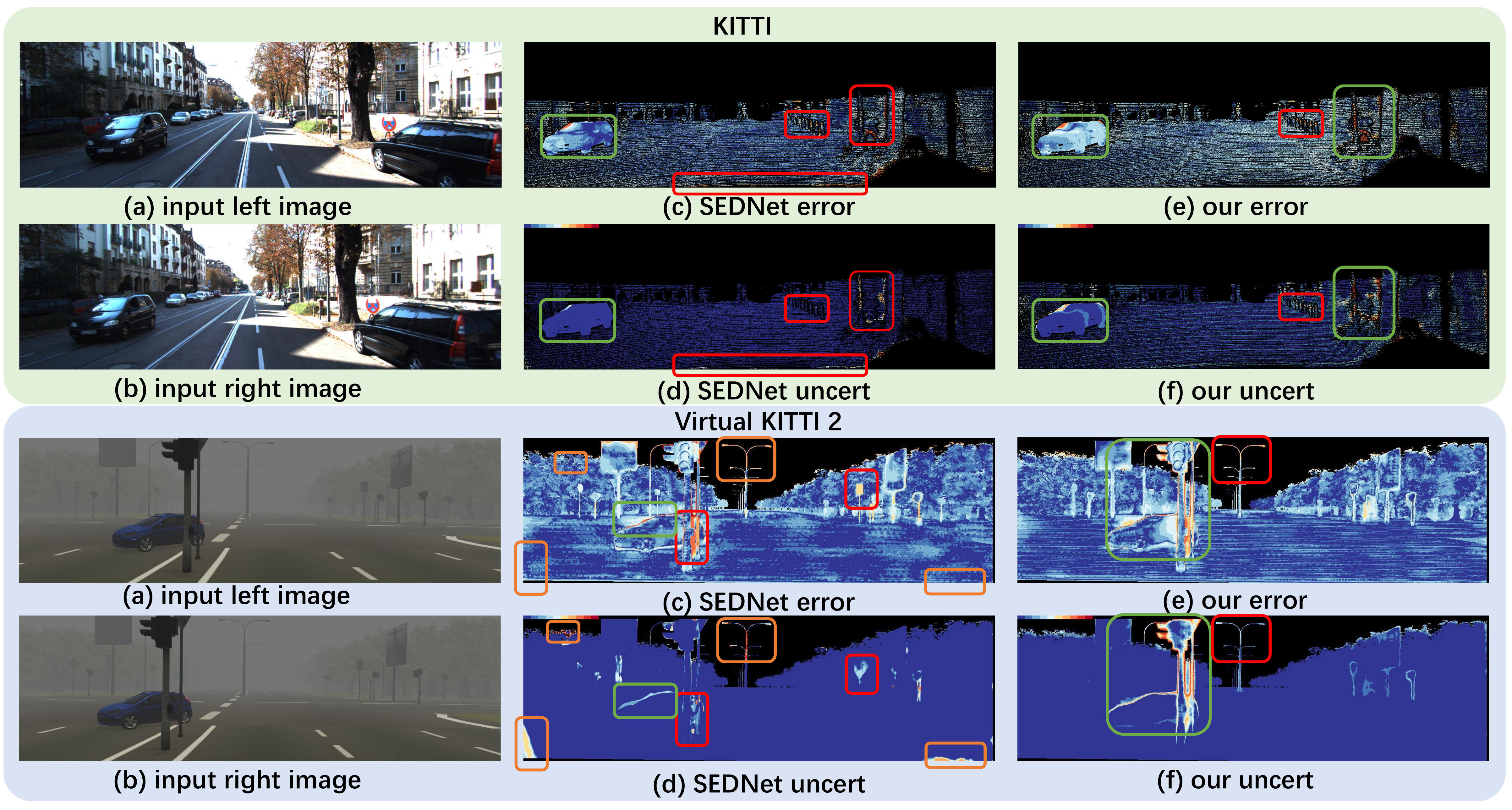}
	\end{center}
	\caption{Experimental results of KITTI~\cite{kitti} and Virtual KITTI~\cite{vk2}. \textcolor{red}{Red} box: Large error, small uncertainty—model fails to predict large errors. \textcolor{orange}{Orange} box: Small error, large uncertainty—model is overly cautious. \textcolor{ForestGreen}{Green} box: uncertainty aligns well with error.}
	\label{fig:exp1}
\end{figure*}

\begin{figure*}[h]
	\begin{center}
		\includegraphics[width=1\linewidth]{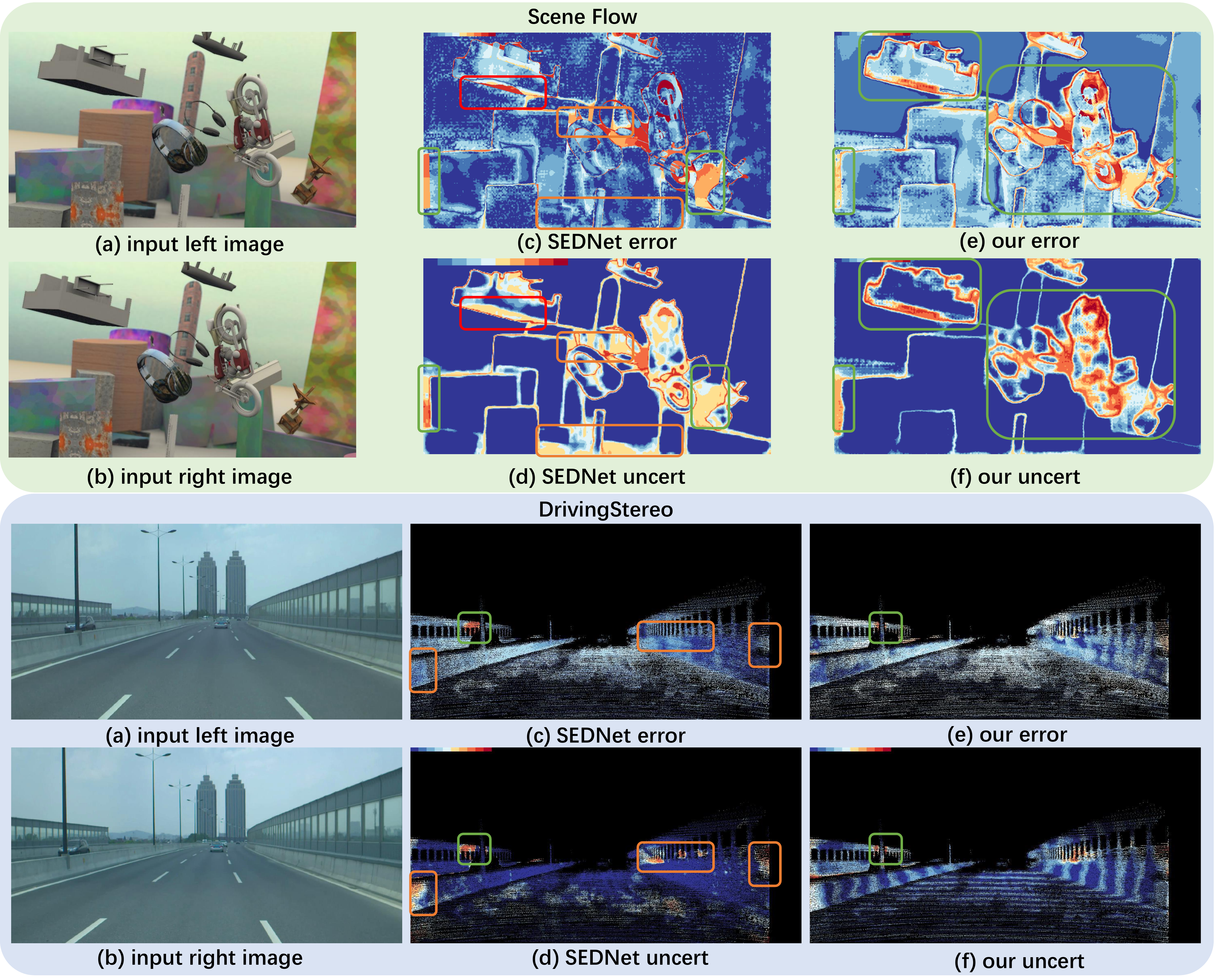}
	\end{center}
	\caption{Experimental results of Scene Flow~\cite{sceneflow} and DrivingStereo~\cite{drivingstereo}. \textcolor{red}{Red} box: Large error, small uncertainty—model fails to predict large errors. \textcolor{orange}{Orange} box: Small error, large uncertainty—model is overly cautious. \textcolor{ForestGreen}{Green} box: uncertainty aligns well with error.}
	\label{fig:exp2}
\end{figure*}

\subsection{Datasets}
We adopt 4 datasets in stereo matching: KITTI~\cite{kitti}, Virtual KITTI~\cite{vk2}, SceneFlow~\cite{sceneflow} and Driving Stereo~\cite{drivingstereo}. 
Experiments on these datasets are shown in Table.~\ref{table-exp}.
As most of these datasets are relatively comprehensive, if all were used for training, the resulting data uncertainty would be much larger than model uncertainty. 
However, in real-world scenarios, the datasets used for training models are only a small portion.
In other words, currently, we cannot train a universally applicable model; the model will always encounter situations in the real world that were not seen in the training set, i.e., situations with high model uncertainty. 
To simulate this situation in reality and to measure the ability of our method to estimate both data uncertainty and model uncertainty, we only use a small portion of the datasets for training.
Experimental results on subsets of dataset are shown in Table.~\ref{table-exp-small}.
Please kindly refer to our supplementary material for dataset split details.

\begin{table*}[h!]
	\caption{Disparity estimation and uncertainty quantification performance of GwcNet~\cite{gwcnet}, SEDNet~\cite{sednet} and our proposed method. Uncertainty estimation performance is measure by AUSE and 95CI. We also measure mean of uncertainty to show if a model can reach a balance between data and model uncertainty. Best results are marked with \textbf{bold} text.}
	\vspace{-0.8em}
	\setlength{\tabcolsep}{4pt} 
	\centering
	
	\begin{tabular}{lcccccccccccccc}
		
		\toprule
		\multirow{2}{*}{Method}& & \multirow{2}{*}{EPE$\downarrow$} & \multicolumn{2}{c}{Uncert.} & \multicolumn{2}{c}{Mean of Uncert.} & & \multirow{2}{*}{EPE$\downarrow$} & \multicolumn{2}{c}{Uncert.} & \multicolumn{2}{c}{Mean of Uncert.} \\
		
		\cmidrule(lr){4-5} \cmidrule(lr){6-7} \cmidrule(lr){10-11}  \cmidrule(lr){12-13} & & & AUSE$\downarrow$ & 95CI $\uparrow$ & Data & Model & & & AUSE$\downarrow$ & 95CI $\uparrow$ & Data & Model \\
		
		\hline
		
		GwcNet+var           &  \multirow{4}{*}{\rotatebox{90}{KITTI}}  & 0.842 & 0.218 &  0.910 & 2.04 & - & \multirow{4}{*}{\rotatebox{90}{VK2}} & 0.641 & 0.283 & 0.935 & 1.99 & - \\
		GwcNet+var+BS    &   & 0.842 & 0.206 &  0.926 & 2.04 & 4.72 &  & 0.641 & 0.265 & 0.947 & 1.99 & 4.04 \\
		SEDNet(UC)           &   & 0.754 & 0.169 &  0.954 & 0.34 & - &  & 0.447 & 0.160 & 0.982 & 0.36 & - \\
		Ours& & \textbf{0.705} & \textbf{0.120} & \textbf{0.970} & 1.97 & 2.13 & & \textbf{0.418} & \textbf{0.113} & \textbf{0.989} & 4.27 & 3.92 \\
		\hline
		GwcNet+var           & \multirow{4}{*}{\rotatebox{90}{SF}} & 0.847 & 0.267 &  0.873 & 8.46 & - & \multirow{4}{*}{\rotatebox{90}{DS}} & 0.700 & 0.188 & 0.934 & 1.44 & - \\
		GwcNet+var+BS           &   & 0.847 & 0.206 &  0.891 & 8.46 & 9.13 &  & 0.613 & 0.164 & 0.942 & 1.44 & 2.36 \\
		SEDNet(UC) &  & 0.567 & 0.134 & 0.968 & 12.70 & - &  & 0.578 & 0.136 & 0.955 & 0.99 & - \\
		Ours&  & \textbf{0.413} & \textbf{0.110} & \textbf{0.987} & 12.47 & 8.42 &  & \textbf{0.541} & \textbf{0.118} & \textbf{0.982} & 2.30 & 1.98 \\
		
		\bottomrule
	\end{tabular}
	\label{table-exp}
\end{table*}

\subsection{Metrics and Baselines}
\textbf{Metrics}.
Our metrics consist of two parts: the model's ability to estimate disparity and its ability to estimate uncertainty. Endpoint error (EPE)~\cite{flownet, Ilg2018UncertaintyEA} is used to measure the difference between the model's estimated disparity and the ground truth. Area Under the Sparsification Error (AUSE)~\cite{ause} between disparity and EPE is used to measure the quality of uncertainty estimation: pixels are sorted by EPE from smallest to largest, and each time 1$\%$ of the pixels with the largest EPE are removed to calculate the Area Under Curve ($\text{AUC}_{\text{gt}}$). Similarly, pixels are sorted by uncertainty from smallest to largest, and $\text{AUC}_{\text{est}}$ is calculated. $\text{AUC}_{\text{est}} - \text{AUC}_{\text{gt}}$ yields AUSE, with smaller values indicating better uncertainty estimation. AUSE is normalized over EPE to eliminate the factor of prediction accuracy.
We also adopt the accuracy under 95\% Confidence Interval (95CI), measuring the proportion of ground truth values falling within the central region of the predicted distribution that contains 95\% of the probability mass.

\textbf{Baselines}.
On softmax output layer of GwcNet~\cite{gwcnet}, we calculate its variance (GwcNet+var), entropy (GwcNet+E) and negative confidence (GwcNet+conf) as data uncertainty. We adopt bootstrapping~\cite{wildbs} (BS) on GwcNet as model uncertainty. 
To calculate total uncertainty, according to definition of uncertainty in Eq.~\ref{eq:regression-uncert}, we directly sum up data and model uncertainty.
SEDNet~\cite{sednet} with GwcNet backbone is compared, training with smooth-l1~\cite{gwcnet},  log-likelihood~\cite{whatuncert}(KG) and divergence~\cite{sednet}(UC) loss.
For all datasets, we consider disparities in the range of 0-64. 

\subsection{Results}
Due to the probabilistic modeling of disparity and ordinal regression loss, our approach pays more attention on the distribution of disparity, resulting in better disparity and uncertainty estimation accuracy in both full dataset and subset training.
Please refer to Table.~\ref{table-exp} and Table.~\ref{table-exp-small} for results. Qualitative results are shown in Fig. ~\ref{fig:exp1} and Fig. ~\ref{fig:exp2}.
The mean uncertainty also shows that out method reaches a balance between data and model uncertainty~\cite{conor}.
Please kindly refer to our supplementary materials for results on Scene Flow and Driving Stereo, and more ablation studies.

\subsection{Training on Small Uncertainty Data}
Pixels with large data uncertainty tend to have more random corresponding labels $Y$ for the same feature $X$ ($Y$ has large variance on condition of $X$).
The ground truth for these pixels is noisy and can mislead the model.
Therefore, we detect data uncertainty and use it to filter out pixels with high data uncertainty.
The filtered dataset can improve stereo matching accuracy. 
We Training the model on Small Uncertainty Data (TSUD)~\cite{elfnet,ucfnet}.
In training, our model predicts data uncertainty simultaneously with disparity, and we use the predicted uncertainty to mask out high uncertainty data. 
TSUD can also be applied to transfer learning~\cite{transfer1,transfer2} and semi-supervised learning~\cite{semi1,semi2,semi3}.
Since our uncertainty quantification comes directly from the estimated PMF, TSUD does not introduce any extra time or computational costs.
We let the model run half of the total number of epochs first, and use TSUD to select 95\% of the data for training in the remaining epochs. For the existing training pipeline, TSUD is plug-and-play, and we did not change the percentage of data selected by TSUD according to different data sets, in order to demonstrate the versatility of TSUD.
The TSUD experimental results shown in Table.~\ref{table-exp-small} shows that it improves disparity and uncertainty estimation performance.

\begin{table*}[h!]
	\caption{Disparity estimation and uncertainty quantification performance by training on a subset of data.}
	\vspace{-0.8em}
	\setlength{\tabcolsep}{4pt} 
	\centering
	
	\begin{tabular}{lcccccccccccccc}
		
		\toprule
		\multirow{2}{*}{Method}& & \multirow{2}{*}{EPE$\downarrow$} & \multicolumn{2}{c}{Uncert.} & \multicolumn{2}{c}{Mean of Uncert.} & & \multirow{2}{*}{EPE$\downarrow$} & \multicolumn{2}{c}{Uncert.} & \multicolumn{2}{c}{Mean of Uncert.} \\
		
		\cmidrule(lr){4-5} \cmidrule(lr){6-7} \cmidrule(lr){10-11}  \cmidrule(lr){12-13} & & & AUSE$\downarrow$ & 95CI $\uparrow$ & Data & Model & & & AUSE$\downarrow$ & 95CI $\uparrow$ & Data & Model \\
		
		\hline
		GwcNet+var & \multirow{10}{*}{\rotatebox{90}{KITTI}} & 1.073 & 0.324 & 0.893 & 2.12 & - & \multirow{10}{*}{\rotatebox{90}{VK2}} & 0.899 & 0.384 & 0.818 & 2.08 & - \\
		GwcNet+var+BS                 & & 1.073 & 0.315 & 0.904 & 2.12 & 4.48 & & 0.899 & 0.381 & 0.907 & 2.08 & 4.20 \\
		GwcNet+E+BS                   & & 1.073 & 0.301 & 0.920 & 1.38 & 4.48 & & 0.899 & 0.379 & 0.910 & 1.24 & 4.20\\
		GwcNet+conf+BS                & & 1.073 & 0.293 & 0.926 & 0.57 & 4.48 & & 0.899 & 0.367 & 0.927 & 0.53 & 4.20 \\
		SEDNet(smooth-l1)             & & 1.066 & 0.302 & 0.921 & 0.99 & -    & & 0.698 & 0.406 & 0.929 & 0.99 & -\\
		SEDNet(KG)                    & & 1.070 & 0.270 & 0.931 & 0.90 & -    & & 0.910 & 0.344 & 0.917 & 0.89 & - \\
		SEDNet(UC)                    & & 1.021 & 0.232 & 0.940 & 0.36 & -    & & 0.890 & 0.311 & 0.952 & 0.82 & - \\
		SEDNet(UC)+TSUD               & & 1.020 & 0.206 & 0.942 & 0.34 & -    & & 0.841 & 0.277 & 0.949 & 0.83 & - \\
		\rowcolor{lightgray} Ours     & & 1.126 & 0.191 & 0.941 & 2.15 & 3.22 & & 0.742 & 0.260 & 0.963 & 4.39 & 4.22\\
		\rowcolor{lightgray} Ours+TSUD& & \textbf{1.006} &\textbf{0.178}& \textbf{0.944} & 2.08 & 2.64 & & \textbf{0.621} &  \textbf{0.239} & \textbf{0.974} & 4.30 & 4.14\\
		
		\hline
		
		GwcNet+var& \multirow{10}{*}{\rotatebox{90}{SF}} &  1.394 & 0.348 & 0.765 & 9.27 & - & \multirow{10}{*}{\rotatebox{90}{DS}} & 1.010 & 0.296 & 0.843 & 1.50 & -\\
		GwcNet+var+BS                 & & 1.394 & 0.291 & 0.781 & 9.27 & 9.41 &  & 1.010 & 0.264 & 0.866 & 1.50 & 2.51\\
		GwcNet+E+BS                   & & 1.394 & 0.304 & 0.787 & 1.48 & 9.41 &  & 1.010 & 0.263 & 0.868 & 1.35 & 2.51\\
		GwcNet+conf+BS                & & 1.394 & 0.295 & 0.793 & 0.58 & 9.41 &  & 1.010 & 0.266 & 0.864 & 0.55 & 2.51 \\
		SEDNet(smooth-l1)             & & 1.241 & 0.316 & 0.702 & 0.99 & -  &  & 0.789 & 0.297 & 0.842 & 0.99 & -\\
		SEDNet(KG)                    & & 1.225 & 0.240 & 0.804 & 1.17 & -  &  & 0.902 & 0.246 & 0.888 & 0.98 & - \\
		SEDNet(UC)                    & & 1.167 & 0.243 & 0.837 & 0.99 & - &  & 0.889  & 0.212 & 0.910 & 0.96 & -\\
		SEDNet(UC)+TSUD               & & 1.148 & 0.231 & 0.847 & 0.99 & - &  & 0.846 & 0.206 & \textbf{0.913} & 0.95 & - \\
		\rowcolor{lightgray} Ours     & & 1.145 & 0.203 & 0.915 & 12.73 & 10.22 & & 0.810  & 0.196 & 0.907 & 2.42 & 2.36\\
		\rowcolor{lightgray} Ours+TSUD& & \textbf{1.135} &  \textbf{0.197} & \textbf{0.920} & 12.70 & 10.07 &  & \textbf{0.725} & \textbf{0.177} & \textbf{0.913} & 2.21 & 2.15 \\
		
		\bottomrule
	\end{tabular}
	\label{table-exp-small}
\end{table*}

\begin{figure*}[h]
	\begin{center}
    \includegraphics[width=1\linewidth]{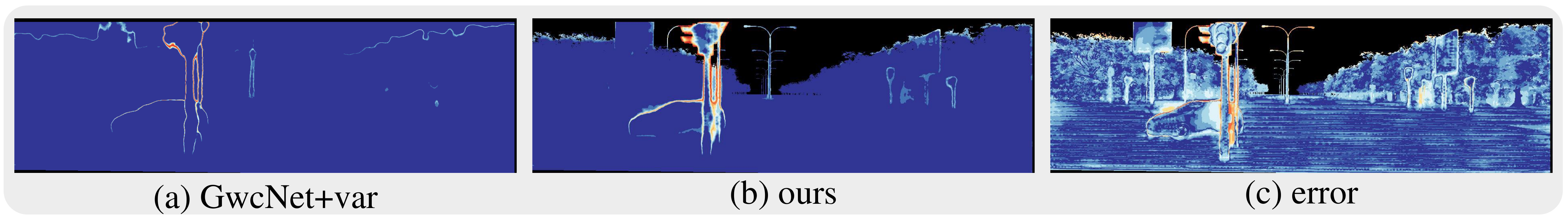}
	\end{center}
 	\caption{(a) GwcNet~\cite{gwcnet} variance as uncertainty. (b) Our uncertainty. (c) Errormap.}
	\label{fig:orloss}
\end{figure*}

\begin{figure}[t!]
	\begin{center}
    \includegraphics[width=1\linewidth]{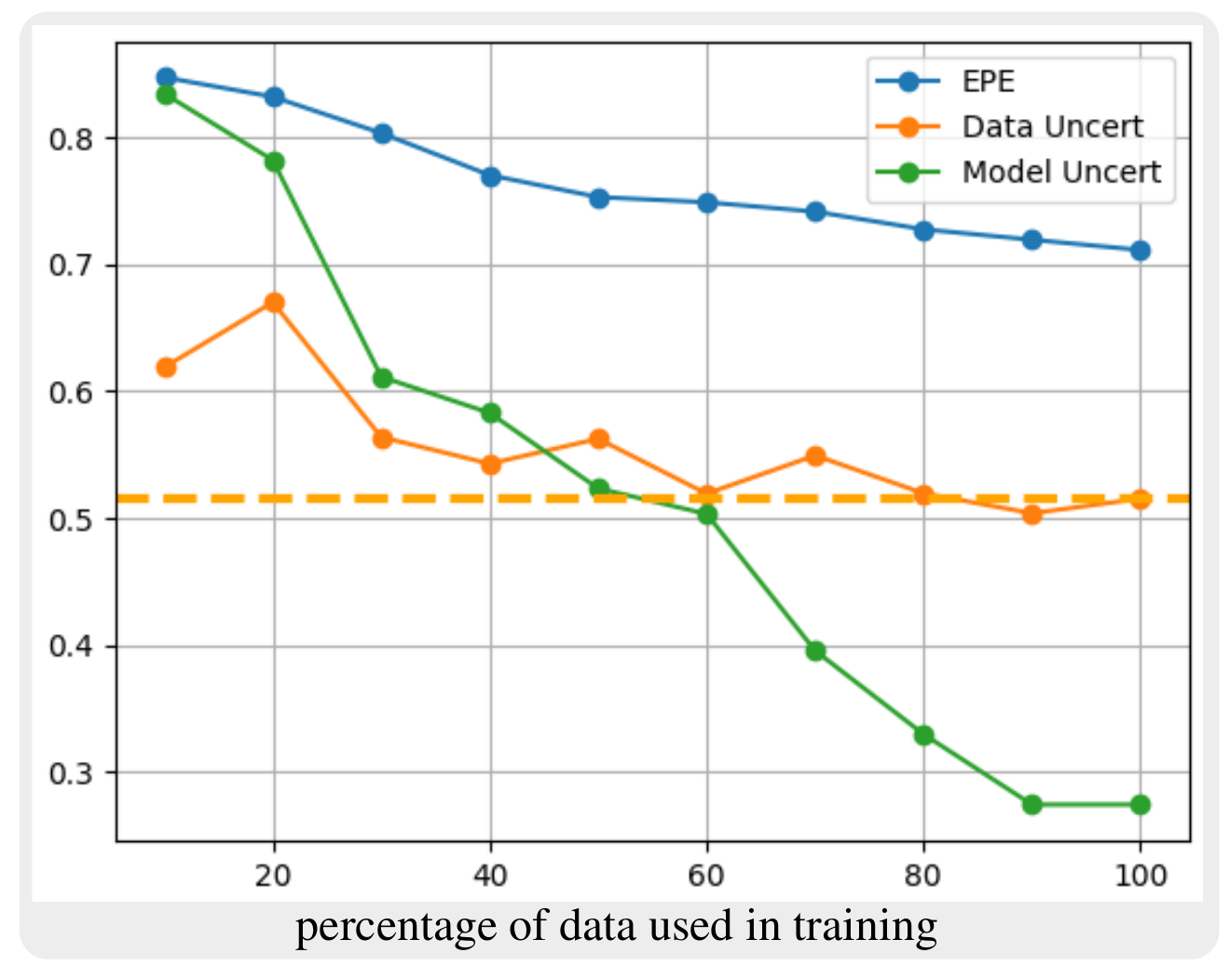}
	\end{center}
 	\caption{Different proportions of KITTI 2015~\cite{kitti} training set is used in training, with model pretrained on Virtual KITTI~\cite{vk2}. We report EPE and mean of data and model uncertainty. EPE and model uncertainty decreases are more training data is used. Our estimation of data uncertainty is consistent and realiable when more than 30$\%$ of training set is used. }
	\label{fig:trainingdata}
\end{figure}

\section{Ablation and Discussion}
\subsection{Ordinal Regression}
Fig.~\ref{fig:orloss} demonstrates the proposed method trained with smooth-l1 loss and ordinal regression loss (OR loss). 
Uncertainty estimation of model trained with OR loss aligns better with error distribution. It provides more detailed information compared with model trained with smooth-l1 loss.
OR loss enables the model to better learn the distribution of disparity, while smooth-l1 loss only provides a prediction value in training.

\subsection{Training Data}
The magnitude of model uncertainty reflects the adequacy of model training. Increasing the training data can reduce model uncertainty, even down to 0. We propose that, to simulate the scenario of insufficient model training in reality, training with a portion of the dataset slightly reduces model performance and increases model uncertainty.
The magnitudes of EPE, data uncertainty, and model uncertainty follow the trend outlined in Fig.~\ref{fig:trainingdata}. Models here are pretrained on 2000 images from Virtual KITTI~\cite{kitti} and trained on KITTI dataset to demonstrate the trend.

\subsection{Data Uncertainty Estimation}
We show ablation studies for disparity bins in ordinal regression.
For disparity within the range [$\alpha, \beta$], we divide it into $K$ bins. 
We uniformly separate the range with bins (Uni),
$B_k = (t_{k-1},t_k]$, where:
\begin{equation}
	t_k = \alpha + k (\beta - \alpha) / K, \; for \, k \in \{ 0,1,...,K \}.
\end{equation}
We also separate disparity bins using in index range (IR):
\begin{equation}
	t_k = exp\left[ log(\alpha) + k log(\beta/\alpha)/K\right], \; for \, k \in \{ 0,1,...,K \}.
\end{equation}
Experimental results on shown in Table.~\ref{table-data-ablation}. We choose Uni with $K=64$ as our model.

\begin{table}[h]
	\caption{Ablation on different disparity bin separations for data uncertainty estimation on KITTI 2015~\cite{kitti}.}
	\setlength{\tabcolsep}{6pt} 
	\centering
	
	\begin{tabular}{lc|cc}
		
		\toprule
		Separation & \# Bins $K$ & AUSE$\downarrow$ & 95CI$\uparrow$ \\
		\hline
		IR & 32 & 0.158 & 0.930 \\
		IR & 64 & 0.145 & 0.937 \\
		Uni & 32 & 0.147 & 0.939 \\
		Uni & 64 & \textbf{0.120} & \textbf{0.970} \\
		\bottomrule
	\end{tabular}
	\label{table-data-ablation}
\end{table}

\subsection{Kernel Regression and Bootstrap}
The Deep Ensemble (bootstrapping)~\cite{ensemble1,ensemble2} require repeated training of the model to estimate uncertainty, thus demanding significant time and computational resources. 
According to~\cite{IsMD}, Monte-Carlo Dropout~\cite{mcdropout1,mcdropout2,mcdropout3} changes the original Bayesian model and its performance would be worse than Deep Ensemble.
We compare the time and model uncertainty estimation performance of KR with Wild Bootstrap (WBS), Multiplier Bootstrap (MBS), and Monte-Carlo Dropout (MCD). Results are reported in Table.~\ref{table-bs}.

\begin{table}[h]
	\caption{Wild Bootstrap~\cite{wildbs}, Multiplier Bootstrap~\cite{multiplierbs}, Monte-Carlo Dropout~\cite{mcdropout1} and Kernel Regression are compared on Scene Flow dataset~\cite{sceneflow}. Data uncertainty is estimated with OR. 2000 images are used in training. Our OR+KR fashion is a little bit worse in model uncertainty estimation compared to bootstrapping, but is significantly faster.}
	\setlength{\tabcolsep}{6pt} 
	\centering
	
	\begin{tabular}{lcccc}
		\toprule
		Method &  WBS~\cite{wildbs} & MBS~\cite{multiplierbs}  & MCD~\cite{mcdropout1} & KR \\
		\hline
		AUSE$\downarrow$  & 10.97 & \textbf{10.93} & 13.66 & 11.84 \\
		95CI$\uparrow$ & 0.925 & \textbf{0.931} &0.904 & 0.915 \\
		Time & 12 hrs & 12 hrs & 4 min & \textbf{1 min}\\
		
		\bottomrule
	\end{tabular}
	\label{table-bs}
\end{table}

\subsection{Model Uncertainty Estimation}
We show experimental results of different kernel functions and bandwidths for model uncertainty estimation. The data uncertainty estimation model here is our model without TSUD.
Results are shown in Table.~\ref{table-model-ablation}.
Finally, we choose RBF kernel for model uncertainty estimation.

\begin{table}[h]
	\caption{Ablation of different kernels and bandwidths in model uncertainty estimation. For polynomial kernel, the bandwidth is (degree, offset) on Scene Flow~\cite{sceneflow}.}
	
	\setlength{\tabcolsep}{6pt} 
	\centering
	
	\begin{tabular}{lc|ccc}
		
		\toprule
		Kernel & Bandwidth & AUSE$\downarrow$ & 95CI$\uparrow$ \\
		\hline
		Polynomial & (1,0) & 0.315 & 0.875 \\
		Polynomial & (1,1) & 0.289 & 0.890 \\ 
		Polynomial & (2,0) & 0.298 & 0.880 \\ 
		Polynomial & (2,1) & 0.276 & 0.904 \\ 
		Epanechnikov & 0.2 & 0.268 & 0.905 \\
		Epanechnikov & 0.5 & 0.275 & 0.910 \\
		Epanechnikov & 1.5 & 0.230 & 0.912 \\
		RBF & 0.5 & 0.275 & 0.902 \\
		RBF & 1 & 0.220 & 0.910 \\
		RBF & 2 & \textbf{0.203} & \textbf{0.915} \\
		RBF & 5 & 0.210 & 0.911 \\
		
		\bottomrule
	\end{tabular}
	\label{table-model-ablation}
\end{table}

\subsection{Training on Small Uncertainty Data}
We adjust the strategy of selecting pixels in the section. First, the model is trained with full dataset to get baseline performance. Then we select 50\% to 99\% data to train the model.
Considering results in Table.~\ref{table-tsud-ablation}, we choose 95\% strategy.
For other data, we use the same strategy instead of fine-tuning the parameters to show the performance of our model without over tricky parameter tuning.

\subsection{Model Uncertainty Matters}
\begin{figure*}[h]
	\begin{center}
		\includegraphics[width=1\linewidth]{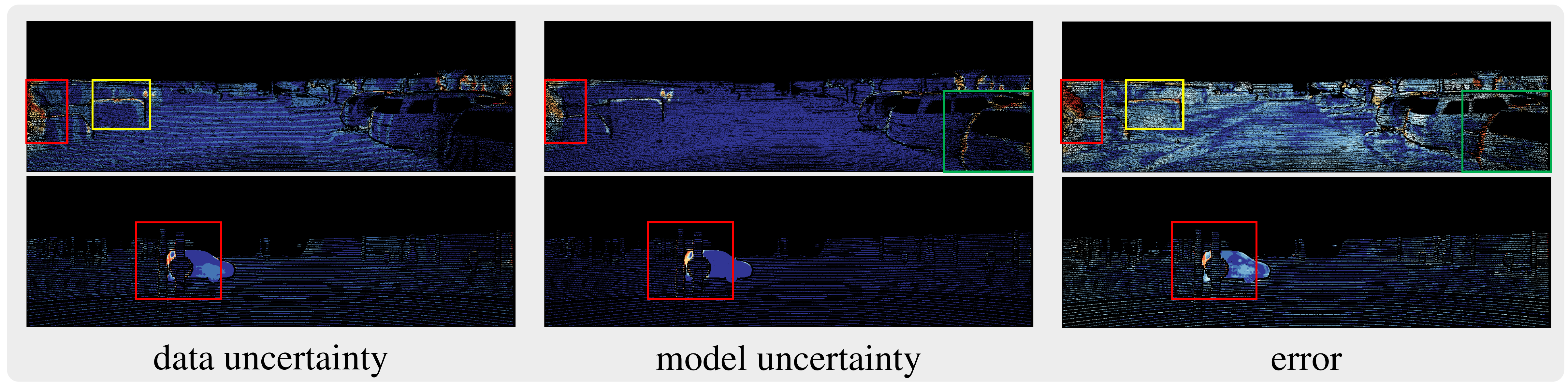}
	\end{center}
	\caption{Model and data uncertainty both explains errors in prediction. Areas where data uncertainty matters most are marked with \textcolor{yellow}{yellow} boxes, while areas where model uncertainty explains error are marked with \textcolor{green}{green} boxes. \textcolor{red}{Red} box indicates that both uncertainty matters.}
	\label{fig:model-data}
\end{figure*}

We use a subset of the dataset for training to assess the model's ability to predict both data and model uncertainty. This training approach has two effects:
\begin{itemize}
	\item Data and model uncertainty are of the same order of magnitude numerically.
	\item We observe that in certain parts of the images, primarily data uncertainty is significant, while in other parts, model uncertainty is predominant. There are also regions where both uncertainties are large. The complementary nature of data and model uncertainty explains the source of errors. Please refer to Fig.~\ref{fig:model-data}. Predicting with data or model uncertainty solely can not fully explain the source of the error.
\end{itemize}

\begin{table}[h]
\caption{Comparisons of data selection strategy in TSUD on KITTI 2015~\cite{kitti}.}

\setlength{\tabcolsep}{6pt} 
\centering

\begin{tabular}{l|cc}

\toprule
 \% of Data & AUSE$\downarrow$ & 95CI$\uparrow$ \\
\hline
50 & 0.169 & 0.939\\
75 & 0.143 & 0.955\\
90 & 0.134 & 0.963\\
95 & \textbf{0.117} & \textbf{0.974} \\
99 & 0.120 & 0.971 \\
100(w/o TSUD) & 0.120 & 0.970 \\

\bottomrule
\end{tabular}
\label{table-tsud-ablation}
\end{table}

\section{Datasets and Broader Impacts}
We adopt four datasets in experiments.

\textbf{KITTI}~\cite{kitti} has two versions, 2012 and 2015, containing 194 and 200 training image pairs sized 1248x384. We use all images from KITTI 2015 and randomly select 10 images from KITTI 2012 for training, with model testing on KITTI 2012.

\textbf{Virtual KITTI}~\cite{vk2} is a synthetic dataset consisting of 21260 image pairs sized 1242x375. We randomly select 2000 pairs for training.

\textbf{SceneFlow}~\cite{sceneflow} images are sized 960x540 and consist of 35454 training and 4370 test stereo pairs. We use the finalpass version and randomly select 2000 images for training.

\textbf{DrivingStereo}~\cite{drivingstereo} contains 174437 training and 7751 test stereo pairs, sized 881x400. 

Apart from training on full dataset, we also train on subset of data to mimic the real world applications.
For KITTI, we choose its 2015 version and a small set of 2012 version as training dataset for subset training. 
For other three datasets, we randomly select 2000 images for training.

It is important to acknowledge that our paper does not explicitly discuss broader impacts in the proposed method, such as fairness or bias. 
Uncertainty itself can be utilized in real-world tasks, such as assisting autonomous vehicles in deciding whether to observe the surrounding environment more cautiously. 
Further research into how our algorithm may interact with other aspects of depth estimation and uncertainty estimation is encouraged.

\section{More Qualitative Results}
\begin{figure*}[h]
	\begin{center}
    \includegraphics[width=1\linewidth]{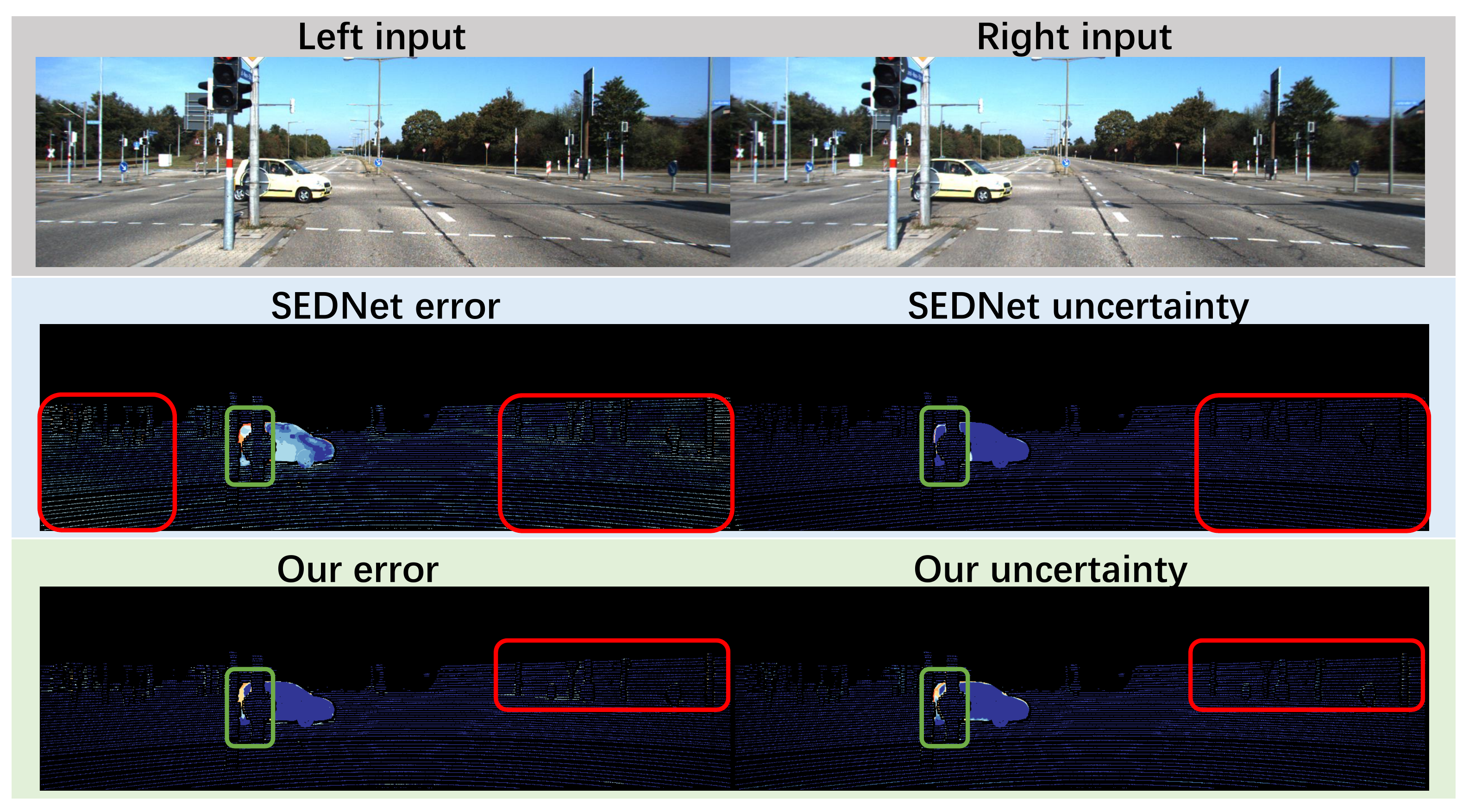}
	\end{center}
 	\caption{Experimental results on KITTI~\cite{kitti}. \textcolor{red}{Red} box: Large error, small uncertainty. \textcolor{ForestGreen}{Green} box: uncertainty aligns well with error.}
	\label{fig:kitti_000043_10}
\end{figure*}

\begin{figure*}[h]
	\begin{center}
    \includegraphics[width=1\linewidth]{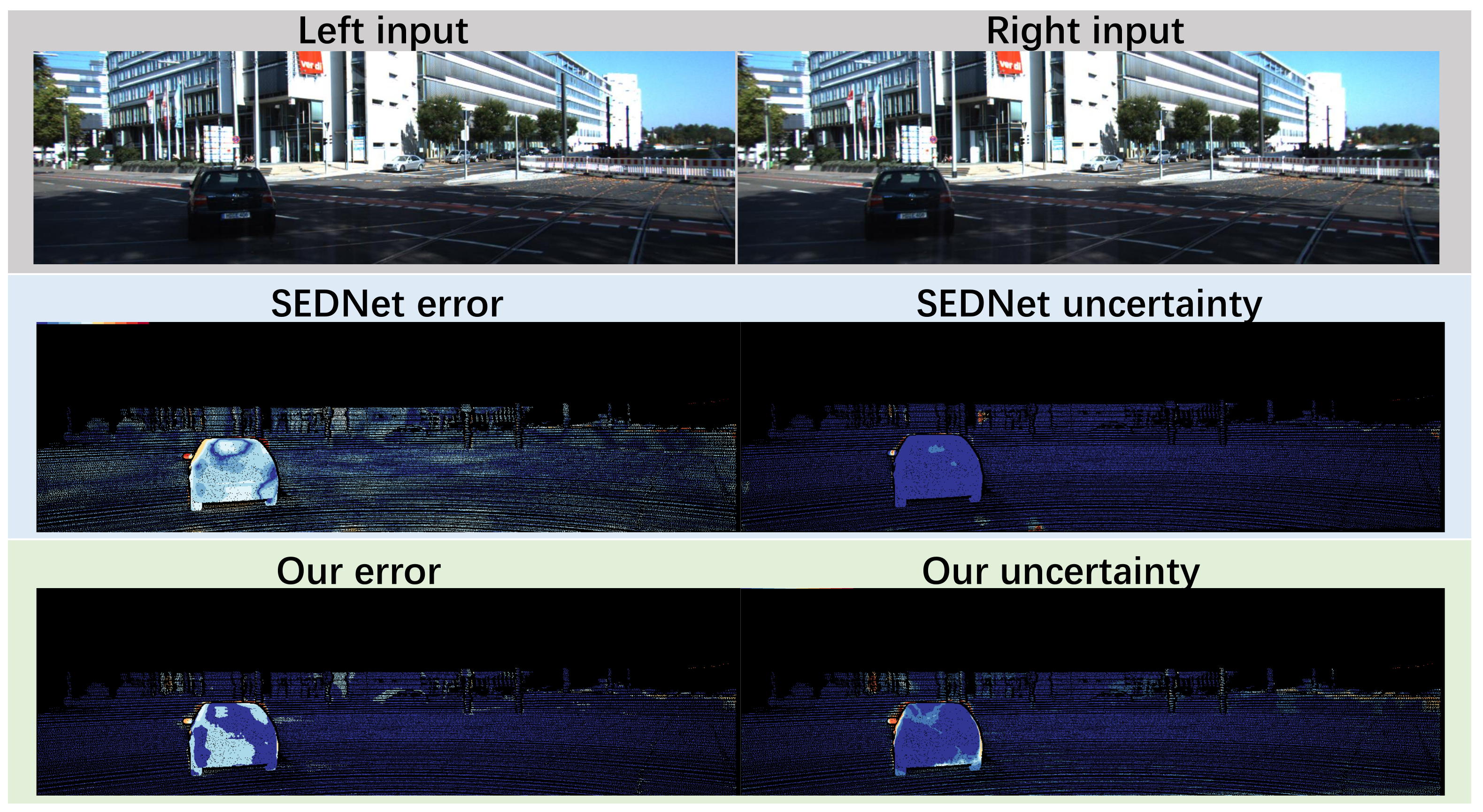}
	\end{center}
 	\caption{Experimental results on KITTI~\cite{kitti}.}
	\label{fig:kitti_000171_10}
\end{figure*}

\begin{figure*}[h]
	\begin{center}
    \includegraphics[width=0.9\linewidth]{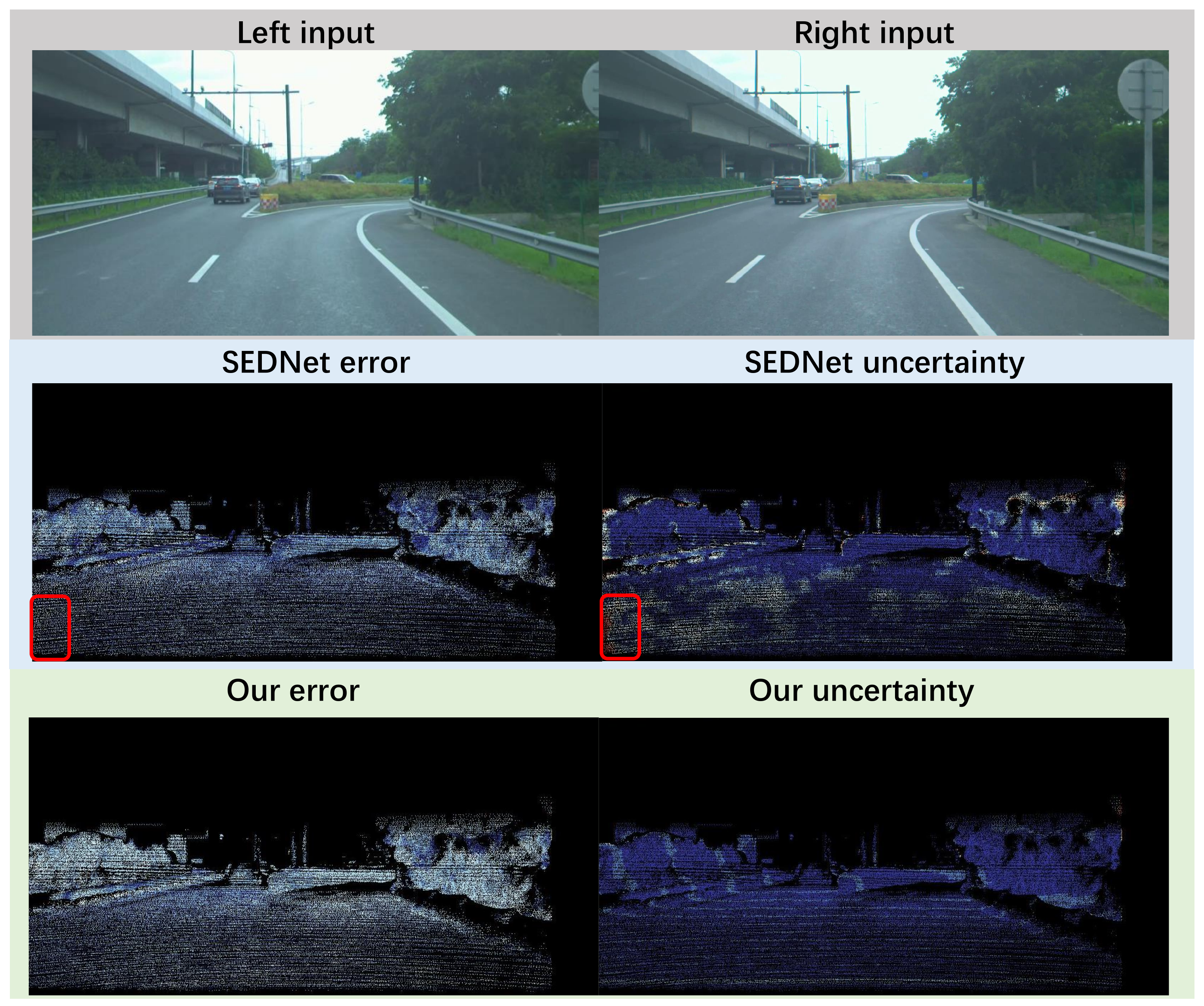}
	\end{center}
 	\caption{Experimental results on DrivingStereo~\cite{drivingstereo}. \textcolor{red}{Red} box: Large error, small uncertainty. }
	\label{fig:ds1}
\end{figure*}

\begin{figure*}[h]
	\begin{center}
    \includegraphics[width=0.9\linewidth]{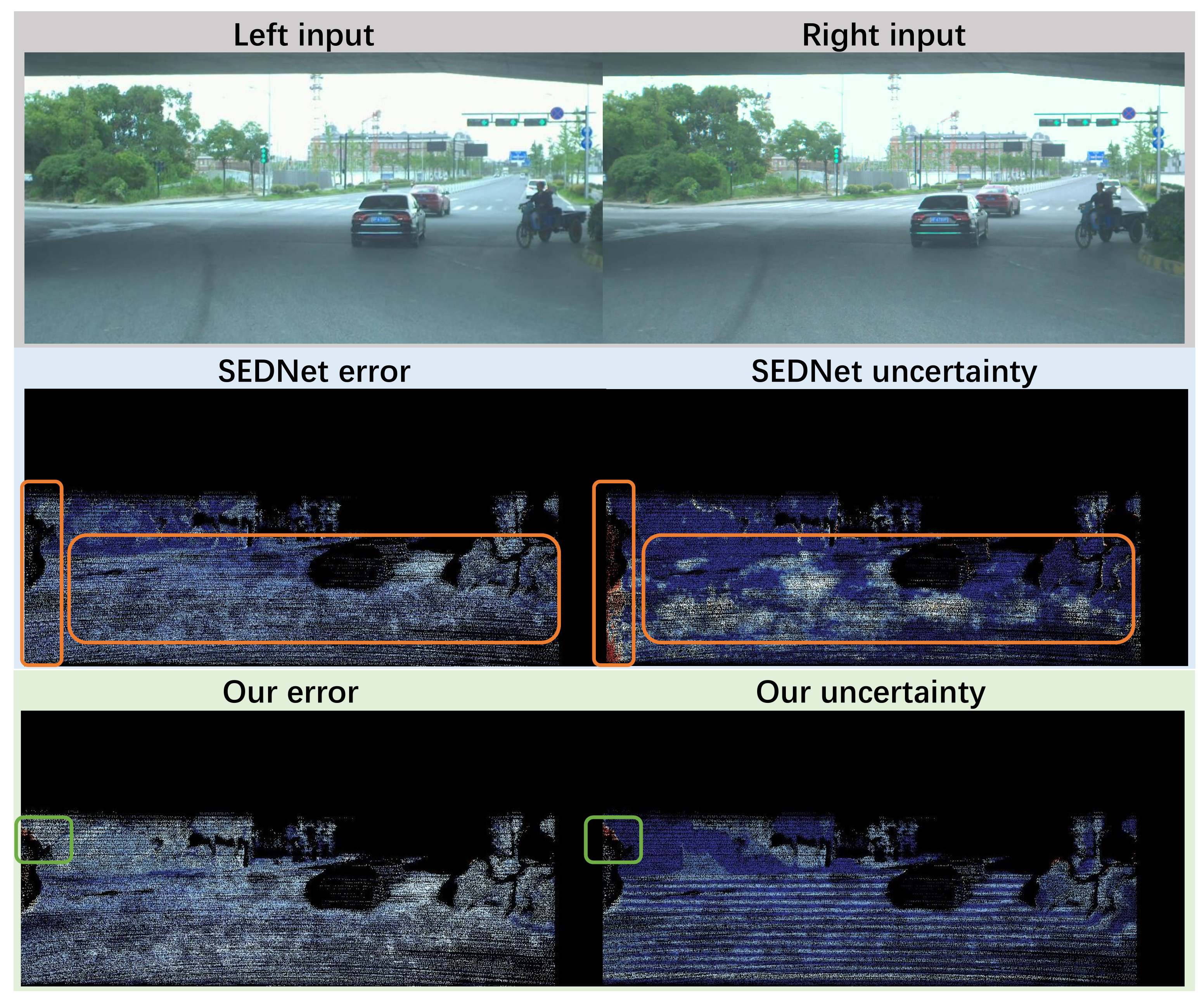}
	\end{center}
 	\caption{Experimental results on DrivingStereo~\cite{drivingstereo}. \textcolor{orange}{Orange} box: Small error, large uncertaint. \textcolor{ForestGreen}{Green} box: uncertainty aligns well with error.}
	\label{fig:ds2}
\end{figure*}

\begin{figure*}[h]
	\begin{center}
    \includegraphics[width=1\linewidth]{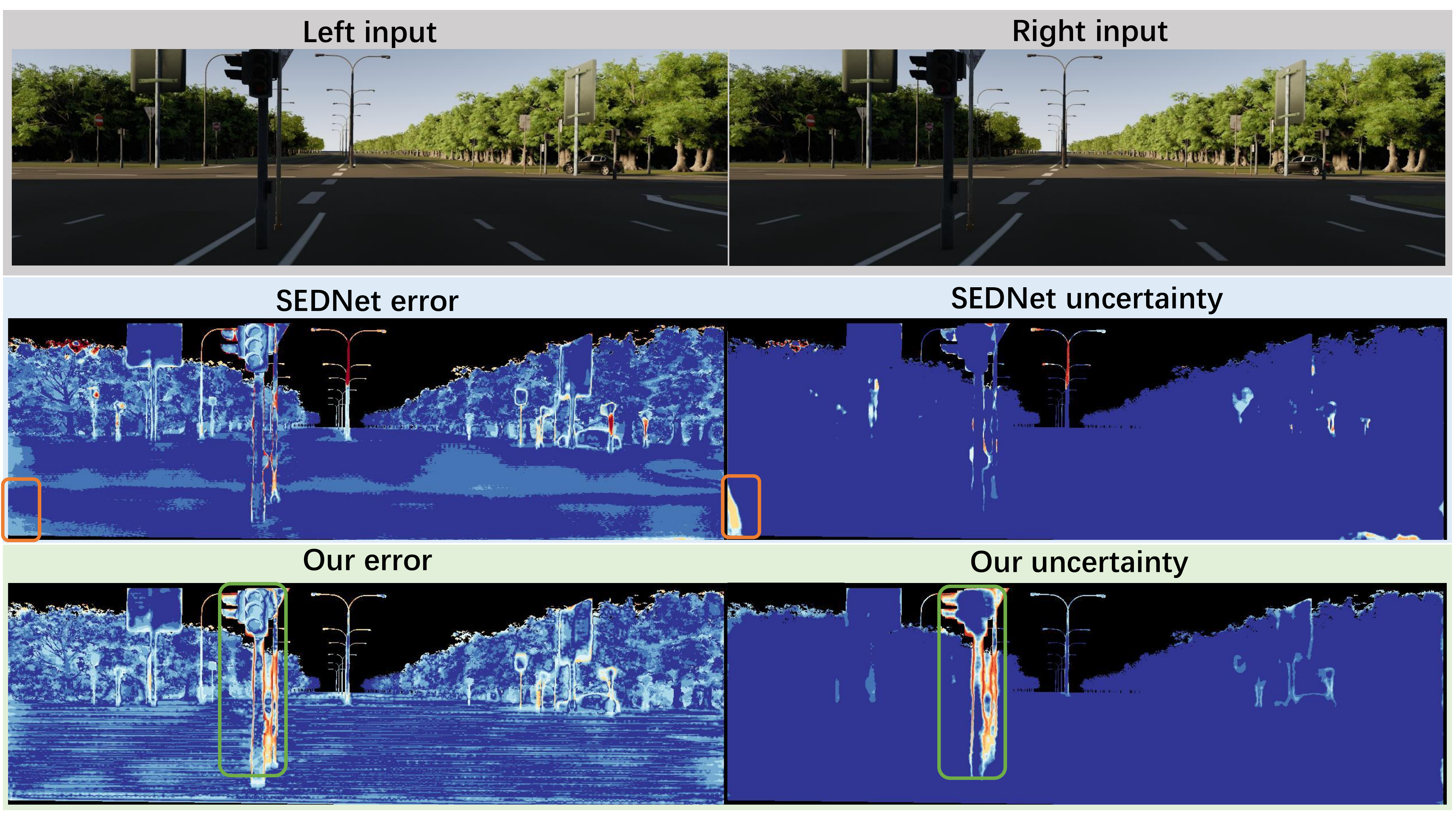}
	\end{center}
 	\caption{Experimental results on Virtual KITTI 2~\cite{vk2}. \textcolor{orange}{Orange} box: Small error, large uncertainty. \textcolor{ForestGreen}{Green} box: uncertainty aligns well with error.}
	\label{fig:vk2-1}
\end{figure*}

\begin{figure*}[h]
	\begin{center}
    \includegraphics[width=1\linewidth]{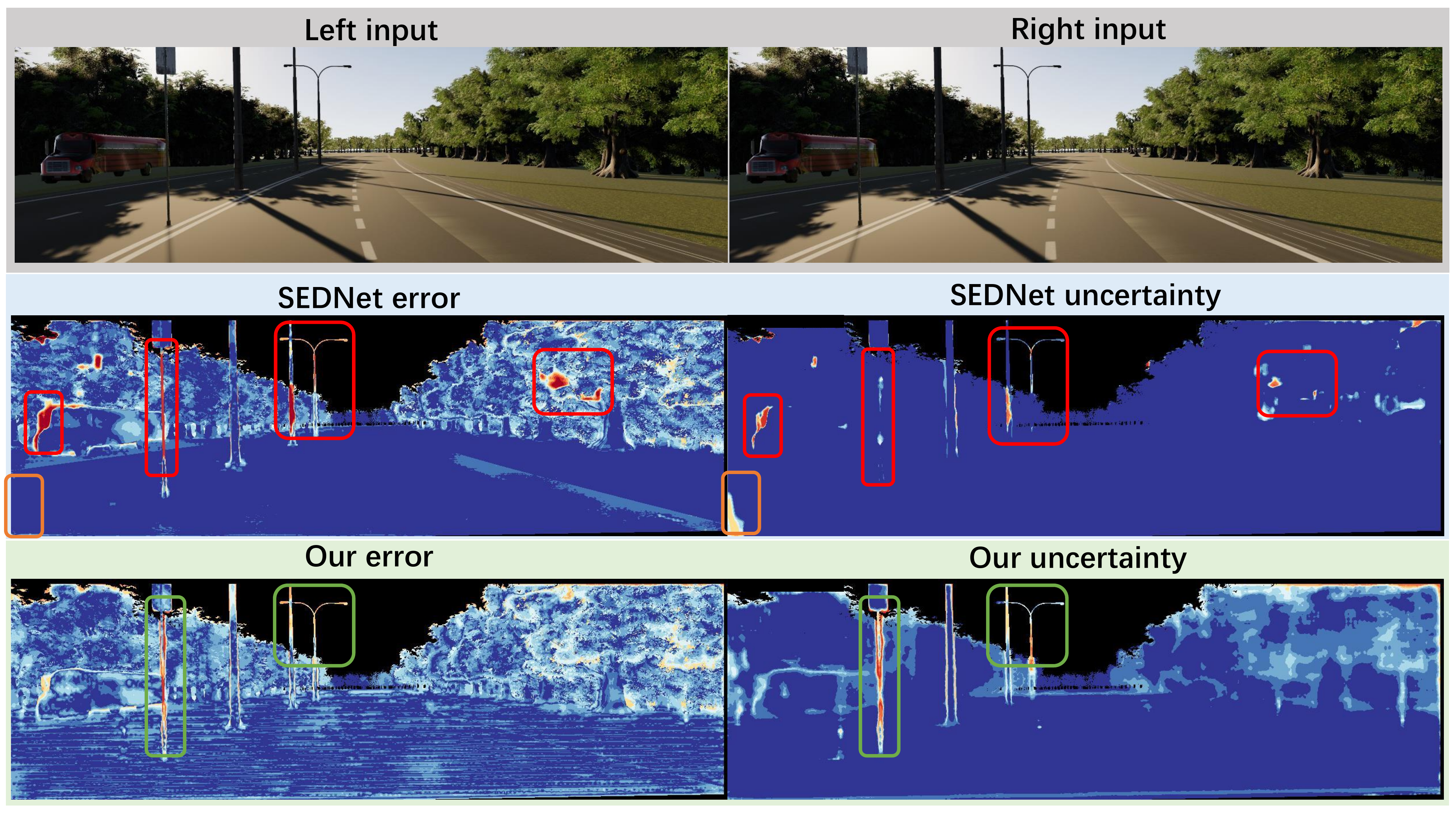}
	\end{center}
 	\caption{Experimental results on Virtual KITTI 2~\cite{vk2}. \textcolor{red}{Red} box: Large error, small uncertainty. \textcolor{orange}{Orange} box: Small error, large uncertainty. \textcolor{ForestGreen}{Green} box: uncertainty aligns well with error.}
	\label{fig:vk2-2}
\end{figure*}

\begin{figure*}[h]
	\begin{center}
    \includegraphics[width=0.8\linewidth]{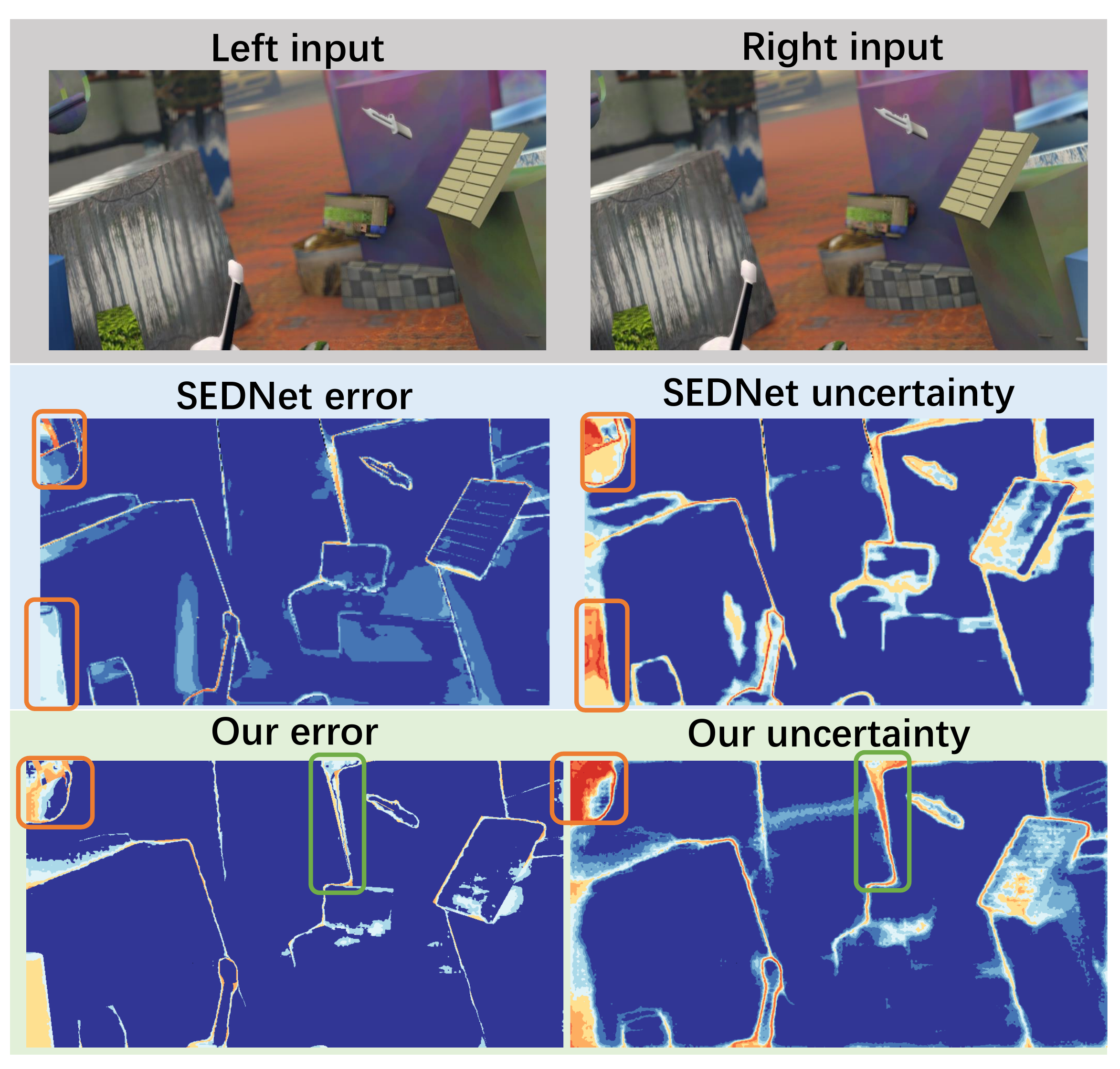}
	\end{center}
 	\caption{Experimental results on Scene Flow~\cite{sceneflow}. \textcolor{orange}{Orange} box: Small error, large uncertaint. \textcolor{ForestGreen}{Green} box: uncertainty aligns well with error.}
	\label{fig:sf1}
\end{figure*}

\begin{figure*}[h]
	\begin{center}
    \includegraphics[width=0.8\linewidth]{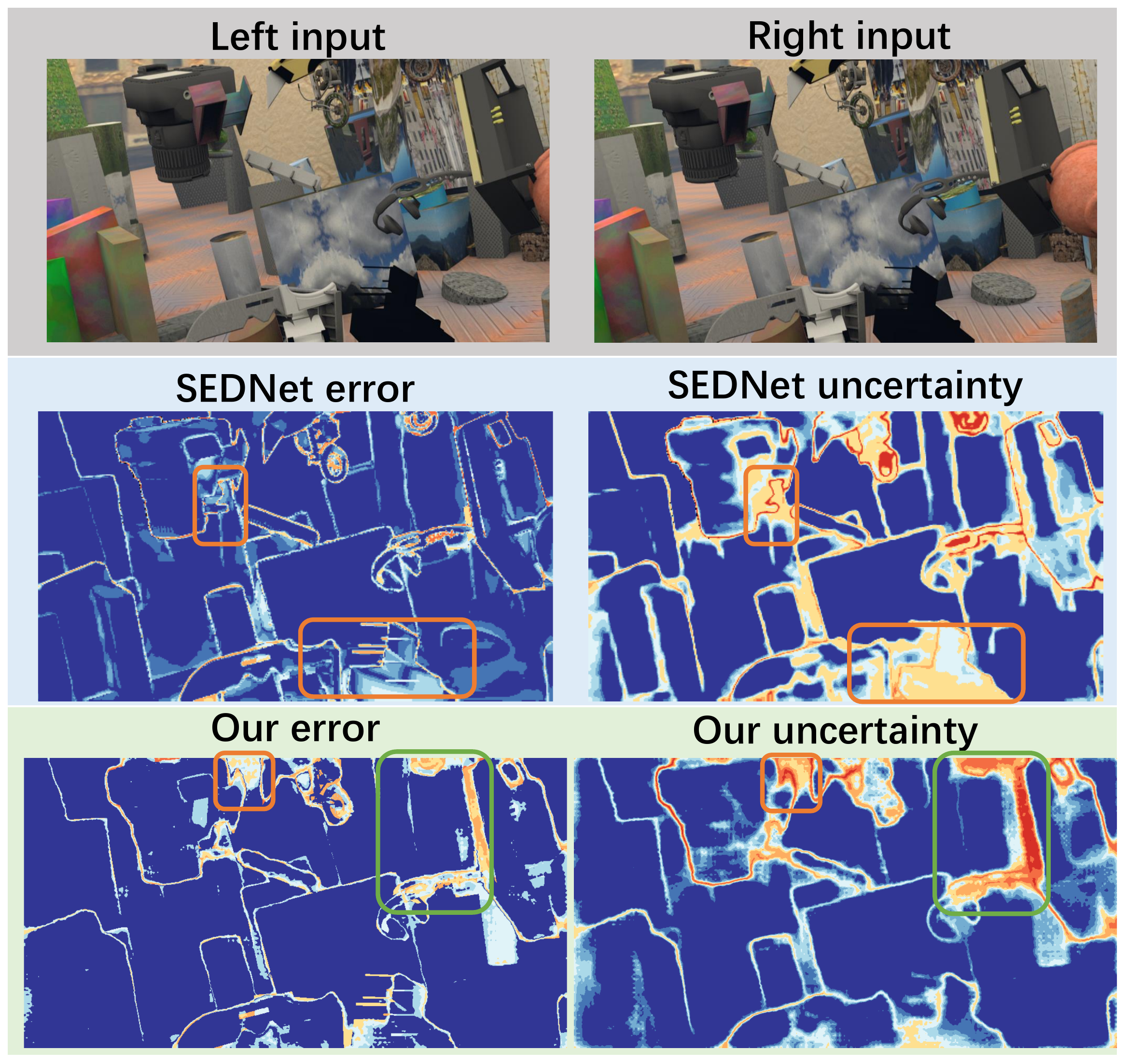}
	\end{center}
 	\caption{Experimental results on Scene Flow~\cite{sceneflow}. \textcolor{orange}{Orange} box: Small error, large uncertaint. \textcolor{ForestGreen}{Green} box: uncertainty aligns well with error.}
	\label{fig:sf2}
\end{figure*}

We show more image results of error and uncertainty pairs in Fig.~\ref{fig:kitti_000043_10},~\ref{fig:kitti_000171_10},~\ref{fig:ds1},~\ref{fig:ds2},~\ref{fig:vk2-1},~\ref{fig:vk2-2},~\ref{fig:sf1} and~\ref{fig:sf2}.

\section{Conclusion}
In this paper, we propose a probabilistic modeling of disparity uncertainty method.
Our approach utilizes ordinal regression for disparity estimation and kernel regression for model uncertainty estimation. 
Experimental results demonstrate that our method can robustly, efficiently and accurately estimating depth and uncertainty. 

\clearpage

\clearpage  

{
    \small
    \bibliographystyle{ieeenat_fullname}
    \bibliography{main}
}

\end{document}